\documentclass[runningheads]{llncs}
\usepackage[T1]{fontenc}
\usepackage{amsmath}
\usepackage{algorithm}
\usepackage[noend]{algpseudocode}
\usepackage{graphicx,verbatim}
\usepackage{color}
\usepackage{hyperref}

\usepackage{xcolor}
\usepackage[normalem]{ulem}
\usepackage{todonotes}
\usepackage{standalone}

\usepackage{placeins}

\begin{document}

\title{Pre to Post-Treatment Glioblastoma MRI Prediction using a Latent Diffusion Model}

\author{Alexandre G. Leclercq\inst{1,2,3}
\and Sébastien Bougleux\inst{1}
\and Noémie N. Moreau\inst{2,3}
\and Alexis Desmonts\inst{2}
\and Romain Hérault\inst{1}
\and Aurélien Corroyer-Dulmont\inst{2,3}
}

\authorrunning{AG. Leclercq et al.}

\institute{Normandie Univ, UNICAEN, ENSICAEN, CNRS, GREYC, Caen, France\\
\email{alexandre.leclercq@unicaen.fr}
\and Artificial Intelligence Department, Centre François Baclesse, Caen, France\\
\and Université de Caen Normandie, CNRS, Normandie Université, ISTCT UMR6030, GIP CYCERON, Caen, France\\
}

    
\maketitle              

\begin{center}
    \small \text{Presented at the Deep Generative Models Workshop of MICCAI (DGM4MICCAI)}
    \small \text{2025, september 2023, Daejeon, South Korea}\footnote{\url{https://doi.org/10.1007/978-3-032-05472-2_18}}
\end{center}

\begin{abstract} Glioblastoma (GBM) is an aggressive primary brain tumor with a median survival of approximately 15 months. In clinical practice, the Stupp protocol serves as the standard first-line treatment. However, patients exhibit highly heterogeneous therapeutic responses which required at least two months before first visual impact can be observed, typically with MRI. Early prediction treatment response is crucial for advancing personalized medicine. Disease Progression Modeling (DPM) aims to capture the trajectory of disease evolution, while Treatment Response Prediction (TRP) focuses on assessing the impact of therapeutic interventions. Whereas most TRP approaches primarly rely on time-series data, we consider the problem of early visual TRP as a slice-to-slice translation model generating post-treatment MRI from a pre-treatment MRI, thus reflecting the tumor evolution. To address this problem we propose a Latent Diffusion Model with a concatenation-based conditioning from the pre-treatment MRI and the tumor localization, and a classifier-free guidance to enhance generation quality using survival information, in particular post-treatment tumor evolution.
Our model were trained and tested on a local dataset consisting of 140 GBM patients collected at Centre François Baclesse. For each patient we collected pre and post T1-Gd MRI, tumor localization manually delineated in the pre-treatment MRI by medical experts, and survival information.


\end{abstract}

\section{Introduction}
Glioblastoma (GBM) is one of the most frequent primary brain tumor, characterized by its aggressive nature and a median survival of approximately 12 to 15 months. In clinical practice, the Stupp protocol \cite{stupp_radiotherapy_2005} is the standard first-line treatment for this pathology. However, treatment responses are highly heterogeneous, with some patients experiencing very short survival times, while others achieve significantly longer survival. Furthermore, the first visual impact of treatment effect on post-treatment MRI (Post MRI) are typically observed no earlier than two months after the beginning of treatments. Thus, it is crucial to find new ways for clinicians to predict treatment efficiency before its initiation, in order to adapt the therapeutic protocols. In particular, Post MRIs are used by radiation oncologist to apply the Response Assessment in Neuro-Oncology (RANO) criteria \cite{wenRANO20Update2023} for evaluating therapeutic response. Therefore, accurately predicting Post MRI outcomes could assist clinicians in anticipating treatment effectiveness and calculating RANO scores prior to treatment initiation. Moreover, obtaining spatial information from the Post MRI scan concerning where the tumour recurrence will appear is also crucial, as this could help to optimise radiotherapy planning.

In recent years, diffusion models \cite{ho2020denoising} have emerged as state-of-the-art methods capable of generating high-quality images, effectively capturing the underlying structure of the image space manifold, and demonstrating strong conditional generation capabilities. More specifically, the Latent Diffusion Model (LDM) \cite{rombach2021highresolution} offers a good trade-off by training the diffusion model in a lower dimensional space that remains perceptually equivalent to the target image space, lowering the computational cost made by the diffusion model by working on lower dimensional space.
Recent studies show that LDM exhibit good performance for multi-domain translation with MRI \cite{kim_adaptative_2024}. However, the source and target domains are always semantically close, and the task does not represent a temporal evolution, as in post-treatment prediction would.

In the literature, Disease Progression Modeling (DPM) is a task aimed at predicting the evolution of a disease based on historical patient data. Specifically, Treatment Response Prediction (TRP) focuses on assessing the impact of therapeutic interventions on disease progression. Various approaches leveraging Deep Learning method have been explored. TRP has been studied from classification perspectives by predicting patient survival categories \cite{han_deep_2020,luckettPredictingSurvivalGlioblastoma2023}.
Other approaches attempt to directly model disease progression using generative models, such as Generative Adversarial Networks (GANs) or, more recently with LDMs. These methods aim to make a temporal prediction of a medical imaging based on a time-series sequence of medical images \cite{ELAZAB2020321,xu_deep_2019,puglisi_enhancing_2024}.
However, to the best of our knowledge, limited research investigated DPM from a single time-point representation of the disease with the use of generative models. In practice, this approach aligns better with clinical needs, as predicting a 4-month Post MRI from a pre-treatment MRI (Pre MRI) enables early decision-making while still allowing the computation of the RANO criteria. This is particularly relevant because treatment effects typically become visible only after approximately four months, and earlier assessments often do not reflect meaningful changes. Moreover, the RANO criteria are based on the 4-month Post MRI, reinforcing the clinical relevance of generating this specific time point. 
In the context of Intracerebral Haemorrhage growth, \cite{ting_intracerebral_2021} proposed to predict the next Non-Contrast Computerized Tomography image and hematoma mask from the previous ones. In this approach, a transformer deforms the input data based on a displacement vector field computed by a U-Net. This approach is still different from our objective as it corresponds to a short time difference (8 to 72 hours). 

In this work, we consider the problem of early visual TRP as a slice-to-slice translation task, generating a Post MRI from a pre-treatment MRI.
We present an LDM with concatenation-based conditioning, incorporating both the Pre MRI and tumor localization. Additionally, we used survival information provided by a classifier to implement classifier-free guidance, directing the generation process to model tumor evolution in accordance with patient survival. The experimental results show the model's ability to generate a post-treatment slice MRI from a pre-treatment slice MRI and the associated tumor localization. Our code is  available at \url{https://github.com/Alexandre-Leclercq/LDM-GBM-prediction}.

\begin{figure}[!t]
\centering
\includegraphics[width=0.86\textwidth]{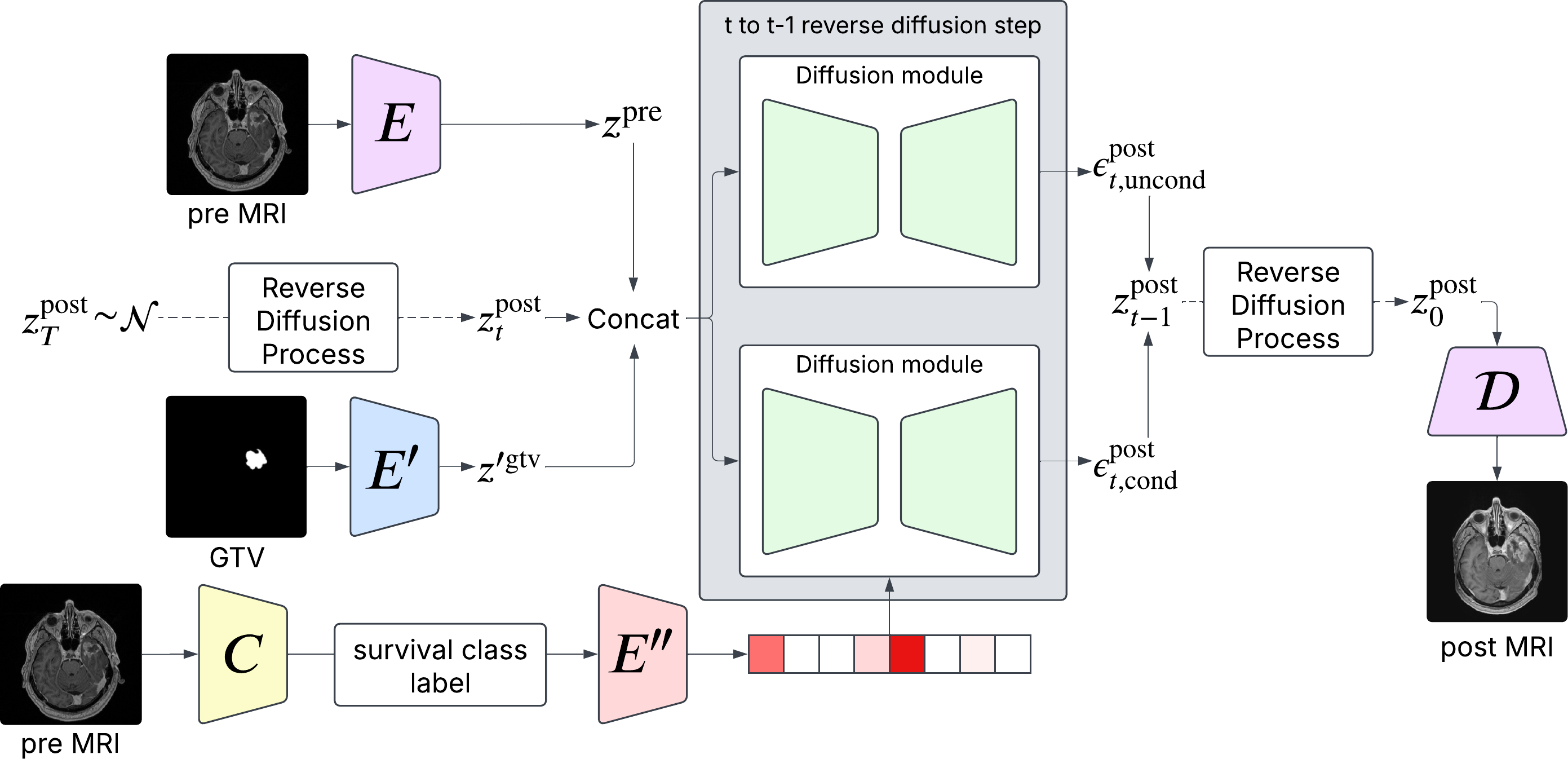}
\caption{Inference pipeline} \label{fig2}
\end{figure}

\section{Method}
Our method aims to generate a 4 month Post MRI slice prediction using the Pre MRI slice, the Gross Tumor Volume (GTV) (corresponding to the contouring of the tumor), and patient survival information (See Fig.~\ref{fig2}). To achieve this, we implemented a 2D-LDM and structured the model training in two major steps. (see Fig.~\ref{fig1}).

\subsubsection{Compression in Latent Space.}
An autoencoder consists of an encoder $E$ and a decoder $\mathcal{D}$, which are learned simultaneously. The encoder compresses an image into a latent space $\mathcal{Z}$ with a lower dimensionality than the pixel space $\mathcal{X}$ while preserving the most relevant information 
for image reconstruction. The decoder reconstructs the image in the pixel space from its latent representation. In particular, Variational Autoencoders (VAEs) learn latent variables approximating a normal distribution.

\begin{figure}[!t]
\centering
\includegraphics[width=0.80\textwidth]{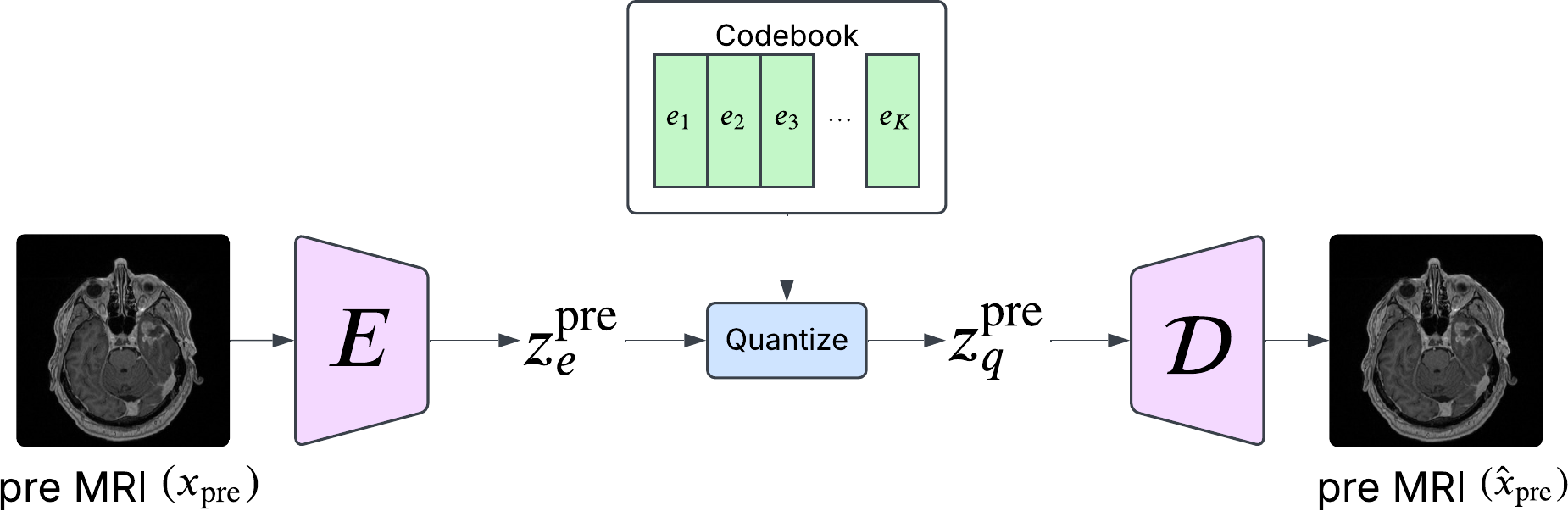}
\caption{VQ-VAE diagram with pre-treatment MRI as example} \label{fig:vq_vae}
\end{figure}

We trained a Vector Quantized-Variational AutoEncoder (VQ-VAE) model \cite{vq_vae_paper} which compresses images into a quantized latent space formed by embedding vectors $e_k$ from a learned codebook (see Fig.~\ref{fig:vq_vae}).
A common latent space $\mathcal{Z}$ was trained for the pre- and Post MRI, as these images are perceptually very similar. 

Separately, another VQ-VAE model was trained to learn a latent space representation $\mathcal{Z}^\prime$ for GTV, 
as GTV differs significantly from MRI. To facilitate concatenation, the dimensionality of the GTV latent space was matched to that of the MRI latent space.

\subsubsection{Latent Diffusion model.}
Diffusion Models are generative models that learn the underlying data distribution of a dataset by denoising data at different steps of a fixed Markov chain of length $T$. A schedule Gaussian noise is added to images $x_0 \sim q(x_0)$ during the forward process at different sequences $t$ of the Markov Chain such that $x_T \sim \mathcal{N}(\textbf{0}, \textbf{I})$. The reverse diffusion process is learned through a denoising model $p_{\theta}(x_{t-1}|x_t)$ (typically a UNet).
These models have achieved state-of-the-art performances in image synthesis \cite{NEURIPS2021_49ad23d1}, generating images with superior quality compared to GANs. Contrary to GANs, diffusion models do not require the use of a discriminator, so they are not subject to mode collapse or adversarial instability.

However, the autoregressive nature of the reverse diffusion process results in longer sampling times compared to other generative models. LDM address this limitation by moving the diffusion model into the latent space of an autoencoder, reducing the computational cost. Moreover, the difference in image generation time between an LDM and a GAN does not have a significant impact in our case, as an LDM still generate images within acceptable timeframes.

\begin{figure}[!t]
\centering
\includegraphics[width=0.86\textwidth]{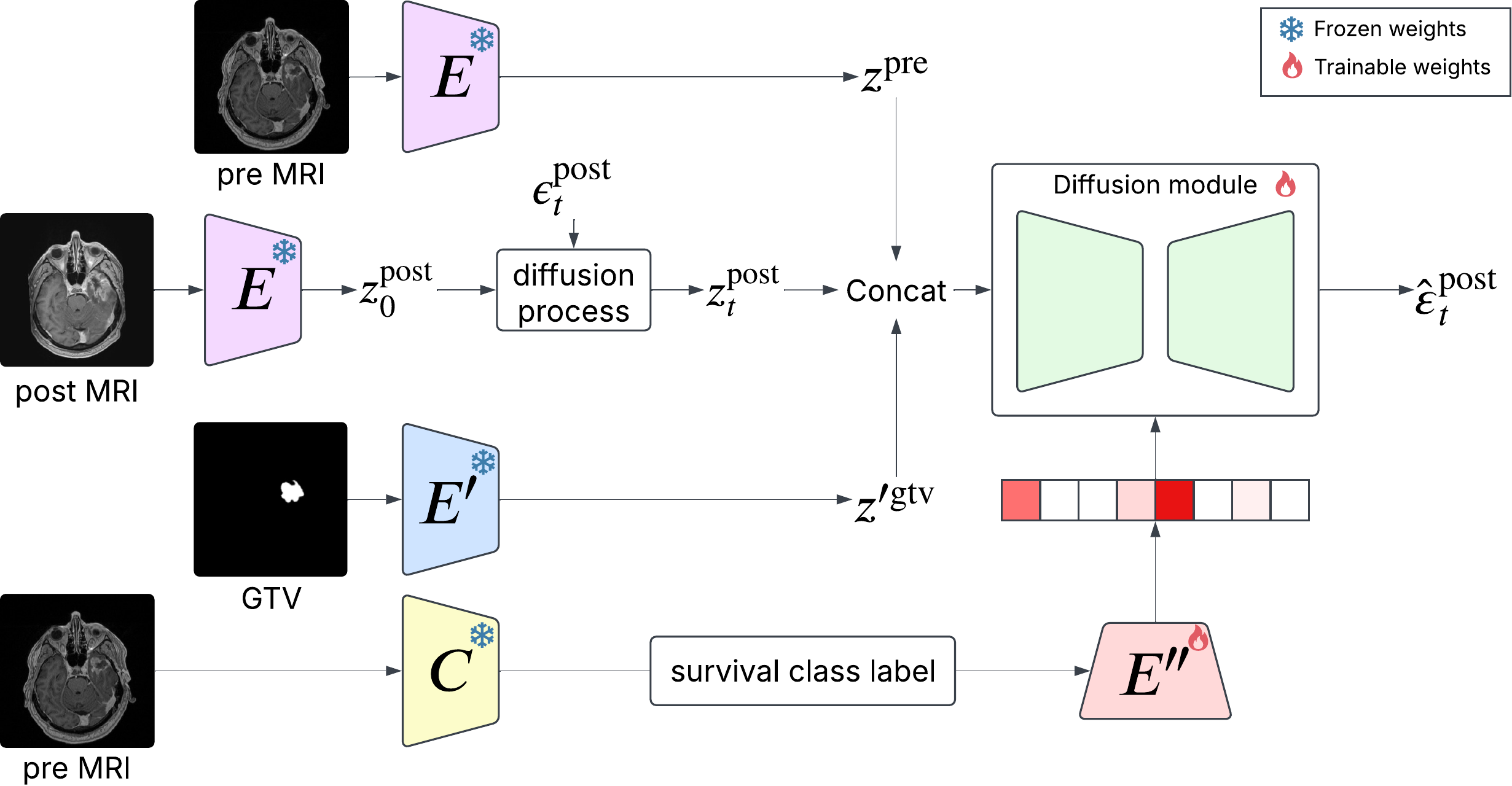}
\caption{Training pipeline} \label{fig1}
\end{figure}

We trained an LDM and conditioned the diffusion process in two ways. First, the model was conditioned by concatenating the Pre MRI latent representation $z^{\textrm{pre}}$ and the GTV latent representation $z'^{\textrm{gtv}}$ with the noisy post-treatment latent representation $z_{t}^{\textrm{post}}$. The Pre MRI provides information about the general structure of the MRI to be generated, while the GTV offers insight into tumor localization, guiding the model's focus.
Additionally, the model was conditioned using patient survival labels as classifier-free guidance \cite{ho2022classifierfreediffusionguidance}. Patient survival labels are predicted from Pre MRI using a pre-trained classifier $C$.

During training, a classification embedding $E''$ was learned and applied at different levels of the UNet ($\epsilon_\theta$) with cross-attention layers.
The goal of the diffusion model during training is to predict the Gaussian noise present in the noisy Post MRI latent representation $\varepsilon_t^{\textrm{post}}$, effectively learning the underlying data distribution of post-treatment MRI. Algorithm~\ref{algo:training} detail training steps, while Fig.~\ref{fig1} summarizes the training pipeline.

\begin{algorithm}[h]
\caption{Training with concatenation-based and class conditioning} \label{algo:training}
\begin{algorithmic}[1]
\State \textbf{repeat}
\State \quad $(z^{\textrm{pre}}, z'^{\textrm{gtv}}, z^{\textrm{post}}, c) \sim q(z^{\textrm{pre}}, z'^{\textrm{gtv}}, z^{\textrm{post}}, c)$
\State \quad $t \sim \textrm{Uniform} (\{1,\dots,T\})$
\State \quad $\epsilon \sim \mathcal{N}(\textbf{0}, \textbf{I})$
\State \quad $z^{\textrm{post}}_t \gets \textrm{ForwardDiffusionProcess}(z^{\textrm{post}}, t, \epsilon)$
\State \quad Take gradient descent step on $\nabla_\theta ||\epsilon - \epsilon_{\theta}(z^{\textrm{pre}} \oplus z'^{\textrm{gtv}} \oplus z^{\textrm{post}}_t, c, t) ||_2^2$
\State \textbf{until} convergence
\end{algorithmic}
\end{algorithm}

During inference (Fig.~\ref{fig2}), we sample $z_T^{\text{post}} \sim\mathcal{N}(0, I)$, which serves as initial states for the reverse diffusion process. Generation is performed using a classifier-free guidance strategy. Specifically, for each step of the reverse diffusion process, two predictions are made using the same model: $\epsilon_{t, \textrm{cond}}^{\textrm{post}}$, the noise present in $z_t^{\textrm{post}}$ conditioned by the survival class; $\epsilon_{t, \textrm{uncond}}^{\textrm{post}}$, the noise present in $z_t^{\textrm{post}}$ without conditioning. To obtain unconditioned class predictions, we provide the diffusion model with a specific class label representing the unconditioned case. Finally, the overall noise estimate $\epsilon_t^{\textrm{post}}$ is computed as a weighted combination of $\epsilon_{t, \textrm{uncond}}^{\textrm{post}}$ and $\epsilon_{t, \textrm{cond}}^{\textrm{post}}$, where a scaling hyperparameter controls the influence of the class-conditioned prediction. Algorithm~\ref{algo:inference} details the generation process.

\begin{algorithm}[h]
\caption{Inference with concatenation-based and class conditioning} \label{algo:inference}
\begin{algorithmic}[1]
\State $(z^{\textrm{pre}}, z^{\textrm{gtv}}, c) \sim q(z^{\textrm{pre}}, z'^{\textrm{gtv}}, c)$
\State $z_T^{\textrm{post}} \sim \mathcal{N}(0,I)$
\For{$t=T,\dots,1$}
\State \textbf{if} $t>1$ \textbf{then} $z \sim \mathcal{N}(\textbf{0},\textbf{I})$ \textbf{else} $z=\textbf{0}$
\State $\epsilon_{t, \textrm{cond}} \gets \epsilon_{\theta}(z^{\textrm{pre}} \oplus z'^{\textrm{gtv}} \oplus z^{\textrm{post}}_t, c, t)$
\State $\epsilon_{t, \textrm{uncond}} \gets \epsilon_{\theta}(z^{\textrm{pre}} \oplus z'^{\textrm{gtv}} \oplus z^{\textrm{post}}_t, t)$
\State $\epsilon_t \gets \epsilon_{t, \textrm{cond}} + scale \times \left(\epsilon_{t, \textrm{uncond}} - \epsilon_{t, \textrm{cond}} \right)$
\State $z_{t-1}^{\textrm{post}} \gets \textrm{ReverseDiffusionProcess}(z_t^{\textrm{post}}, \epsilon_t)$
\EndFor
\State \textbf{return} $z_0^{\textrm{post}}$
\end{algorithmic}
\end{algorithm}

\section{Experiments}
\subsubsection{Dataset.}
Our model was trained on a local dataset comprising 140 patients treated for a glioblastoma (GBM) at Centre François Baclesse between January 2018 and December 2023. This retrospective study was approved by the local Institutional Review Board and conducted in accordance with the Declaration of Helsinki and MR-004 guidelines. Informed consent was obtained from all participants. For each patient, we collected: a pre-treatment T1-Gd MRI acquired after the surgical resection but before the beginning of radiotherapy; a post-treatment T1-Gd MRI acquired 4 months after the beginning of the radiotherapy; the associated Gross Tumor Volume (GTV) of the Pre MRI, manually segmented by radiation oncologists; and, the survival outcomes, defined as the the number of days between the date of death and the date of beginning of treatment. These survival times were decomposed into 2-class by splitting at the median, and into 4-class by splitting at the 25th, 50th and 75th percentiles as thresholds.
These data were selected as they are routinely collected in clinical practice. Each image has a resolution of 256×256 pixels. Additionally, for each patient, we only included MRI slices with a tumor. This yielded approximately 40 slices per patient, resulting in a dataset of 6\,059 slices.
The dataset was split at the patient level using an 80\%/10\%/10\% ratio for the training, validation, and testing sets, respectively. The same data partitioning strategy was applied across all models, ensuring that the test sets for VQ-VAE and LDM were identical. Finally, we performed data augmentation using random horizontal flipping to leverage the brain's axial symmetry.

\subsubsection{Implementation Details.}
Models were implemented using PyTorch and Torch Lightning.
Our LDM implementation is based on Stable Diffusion \cite{rombach2021highresolution}. All models were trained on two NVIDIA RTX A6000 ($2\times48$ GB VRAM). For the LDM training, we fixed a batch size of 30 and used the AdamW optimizer with a learning rate of $2\times10^{-6}$. The training process employed 1\,000 timesteps  and used DDIM sampling \cite{song2020denoising} with 200 steps. Each model was trained for 200 epochs, with the best-performing epochs saved based on the evaluation results for the validation set. For classifier-free guided diffusion, we used a guidance scale of 10.

\subsubsection{Evaluation Metrics.}
To evaluate our model on the test set with quantitative metrics, we computed PSNR, SSIM, LPIPS and MSE between source and generated images. Since our approach aimed to predict tumor evolution, we also computed PSNR, SSIM, and MSE locally to the GTV. 
Results are given as the mean and standard deviation of these metrics for the same epoch.

\section{Results and Discussion}
We trained a VQ-VAE on the pre- and post-treatment MRI and obtained the following metrics on the test set: a PSNR of 42.08; an SSIM of 0.988; and a LPIPS of 0.008. We then used this VQ-VAE for the following LDM implementations.

\begin{table}[!t]
    \centering
    \caption{Comparison of different models based on various metrics.}\label{tab:metric}
    
        \begin{tabular}{lccccccc}
            \hline
            LDM & PSNR$\uparrow$  & PSNR$_{local}\uparrow$ & SSIM$\uparrow$ & SSIM$_{local}\uparrow$ & LPIPS$\downarrow$ & MSE$\downarrow$ & MSE$_{local}\downarrow$ \\
            \hline
            no class & $24.65\pm1.74$ & $18.29\pm3.41$ & $0.826\pm0.044$ & $0.306\pm0.150$ & $0.076\pm0.020$ & $0.015\pm0.006$ & $0.078\pm0.060$  \\
            pred 2-class & $\textbf{24.88}\pm\textbf{1.68}$ & $18.48\pm3.66$ & $\textbf{0.830}\pm\textbf{0.043}$ & $0.303\pm0.161$ & $\textbf{0.073}\pm\textbf{0.020}$ & $\textbf{0.014}\pm\textbf{0.005}$ & $0.077\pm0.060$ \\\hline
            2-class  & $24.71\pm1.65$ & $18.57\pm3.51$ & $0.827\pm0.041$ & $0.312\pm0.158$ & $0.074\pm0.019$ & $\textbf{0.014}\pm\textbf{0.005}$ & $0.074\pm0.059$ \\
            4-class & $24.62\pm1.62$ & $\textbf{18.66}\pm\textbf{3.38}$ & $0.825\pm0.041$ & $\textbf{0.313}\pm\textbf{0.156}$ & $0.076\pm0.019$ & $0.015\pm0.006$ & $\textbf{0.071}\pm\textbf{0.052}$ \\
            \hline
        \end{tabular}

\end{table}

We trained one LDM with classifier-free guidance using a pretrained ResNet-based classifier with an accuracy of 83.78\% learned on the same patient cohort. This represents a real life scenario with a post-treatment prediction only realized from a pre-treatment MRI and a GTV as input \cite{moreau2025aidriven}.
In addition, we also trained 3 other model configurations: no classifier; 2-class and 4-class derived from ground truth survival. These implementations were design to study the impact of classifier-free guidance on model performance, as well as the effect of varying the outcome representation.
The metric results for these cases are provided in Table~\ref{tab:metric}.
The LDM conditioned on a 2-class classifier (denoted as pred 2-class) achieved overall better results, while the LDM conditioned on a 4-class label categorization performed best on local metrics. These results suggest that while class conditioning provides a modest performance improvement, the granularity of the outcome representation has limited impact.

\begin{figure}[!t]
\includegraphics[width=\textwidth]{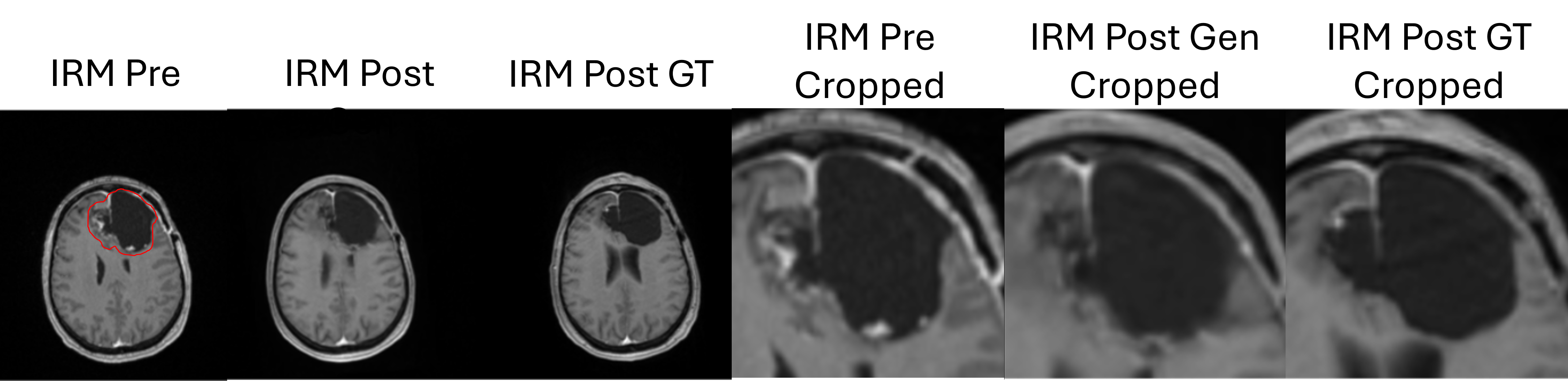}\\
\includegraphics[width=\textwidth]{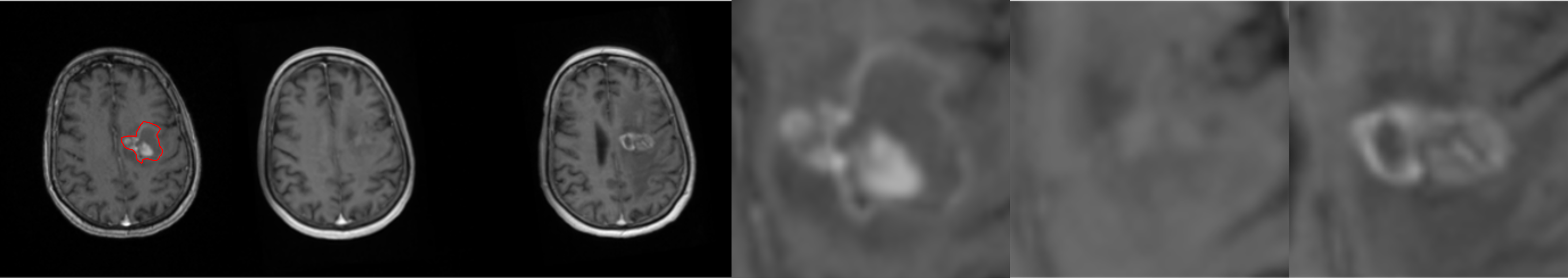}\\
\includegraphics[width=\textwidth]{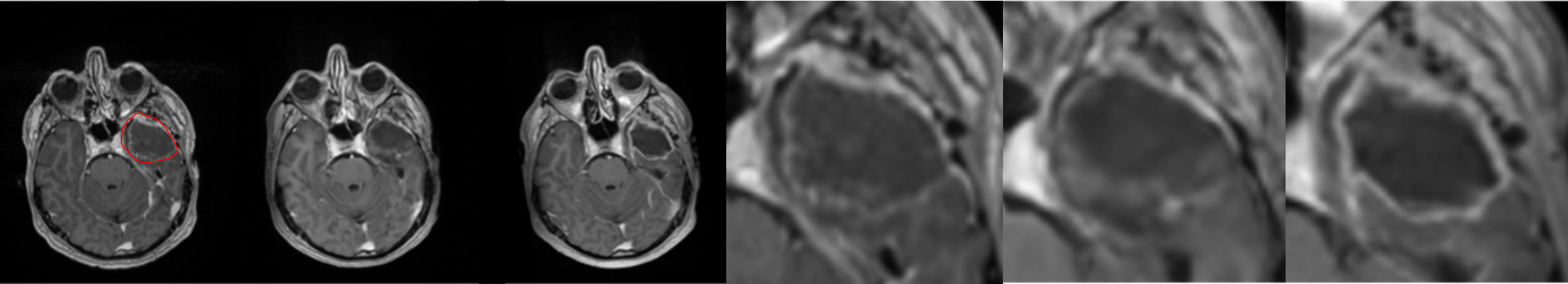}\\
\includegraphics[width=\textwidth]{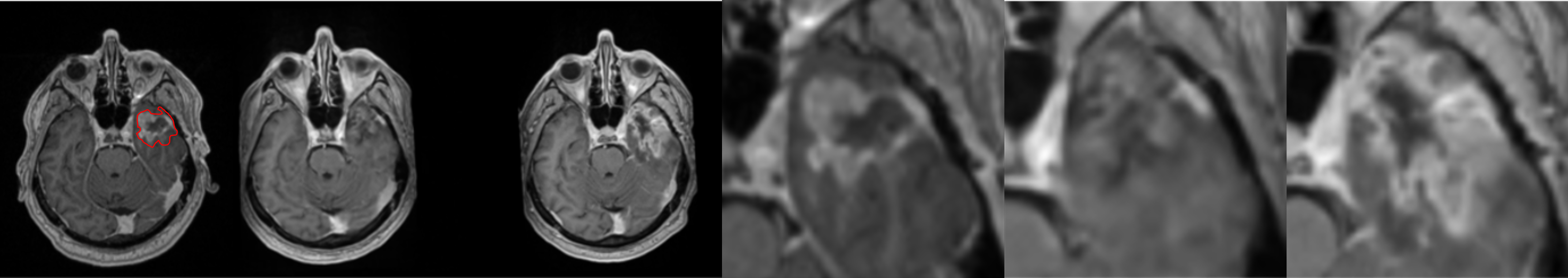}\\
\includegraphics[width=\textwidth]{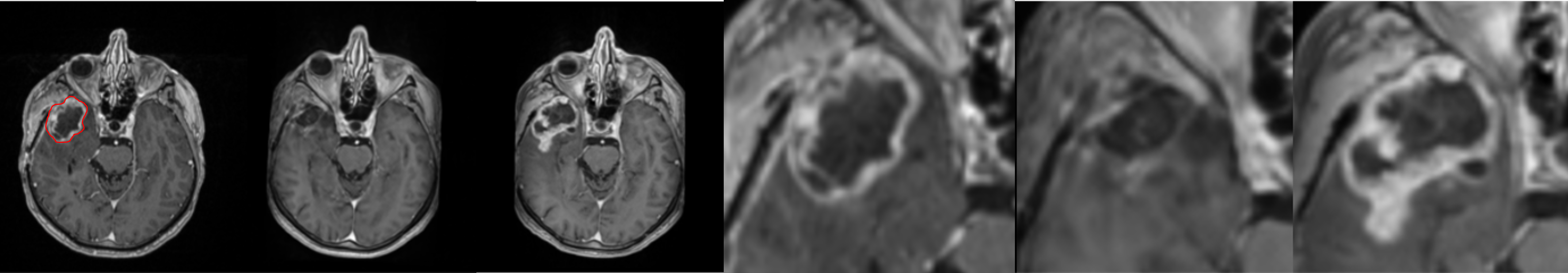}\\
\includegraphics[width=\textwidth]{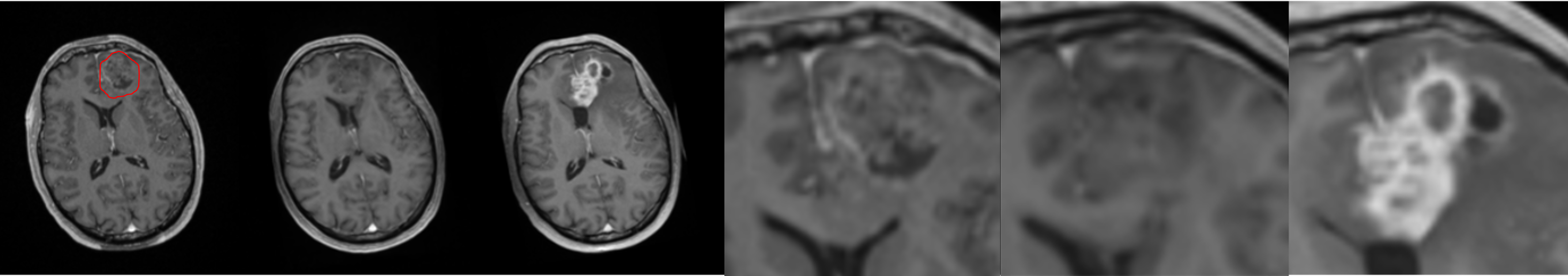}
\caption{(Best viewed in color) Post MRI predictions with the pred 2-class LDM model on the test set, ordered by descending local SSIM (ranging from 0.71 to 0.02). 
The first three columns are the Pre MRI with GTV in red, the predicted Post MRI, and the ground truth Post MRI. Similarly, the last three columns contain the same elements, but centered on the tumor crop.
} \label{fig3}
\end{figure}

A qualitative evaluation of the model was conducted by a medical physicist, who reviewed one generated post-treatment T1-Gd MRI slice per model for each of the 14 patients in the test set. The physicist assigned a score from 0 to 10, reflecting the perceived likelihood of the generated tumor being realistic. The results were generally consistent across all models. The four models (no class, pred 2-class, 2-class and 4-class) achieved an average score of 6.5, 6.0, 7.0 and 7.0, respectively, with standard deviation 1.02, 0.80, 0.80 and 1.02. Minimum scores were 4, 4, 6 and 4, while maximum scores were 8, 7, 8 and 8. 

To further illustrate the model's performance we retained only the pred 2-class model, as it reflects a realistic use case. We present several generated examples in Fig.~\ref{fig3}.
The first row corresponds to the highest local SSIM score (0.71), depicts a patient who underwent total resection, where tumor progression is minimal. We observe that the model fails to predict the hyper-intensity in the periphery of the tumor (T1-enhancement) in comparison to the ground truth.
In contrast, the last row corresponds to the lowest SSIM score (0.02), making it an edge case.
We observe that the model generally failed to predict T1-enhancement localized to the tumor area; however, it succeeded in predicting the necrotic area.

\section{Conclusion}

In this study, we proposed an LDM implementation for early visual TRP in glioblastoma patients. Predicting a post-treatment imaging from a single time-point representation of the disease appeared to be a highly challenging task. In this work, we provided a first proposition to address this problem. 
We investigated the use of classifier-free guidance to guide the generative process regarding the survival outcomes. Our evaluation shows that the use of a classifier have no significant impact on the Post MRI prediction.
We identify several challenges that should be addressed in future work. Firstly, the evaluation metrics currently lack correlation with the task objectives. Specific investigation should be done to identify an evaluation method that aligns with the clinical goals. Secondly, the current model lacks 3D context due to computational constraints. 
We plan to address these limitations in further studies.

\begin{credits}
\subsubsection{\ackname}
This study is supported by the Normandy Region of France and the Centre François Baclesse.

\subsubsection{\discintname}
The authors have no competing interests to declare that are relevant to the content of this article.

\end{credits}

\FloatBarrier


%
%
%
\bibliographystyle{splncs04}
\bibliography{paper}
\end{document}